\documentclass[12pt]{article}
\usepackage{latexsym,amsmath,amsthm,amssymb,amscd,xy,epsfig,graphicx,epstopdf}
\usepackage[round]{natbib}
\usepackage{apalike}
\usepackage{mathptmx}
\usepackage[mathscr]{euscript}
\usepackage{bigints}
\usepackage{caption}
\usepackage{float}
\newfloat{algorithm}{t}{lop}
\usepackage{color}
\usepackage{fullpage}
\usepackage[title]{appendix}
\usepackage[pdftex]{hyperref}
\usepackage{algorithm}
\usepackage{algpseudocode}

\newtheorem{Pro}{Proposition}

\date{}\title{Bayesian Optimisation for Sequential Experimental Design with Applications in Additive Manufacturing}
\usepackage{authblk}
\author{\normalsize{$^{*}$Mimi Zhang$^{1, 4}$, Andrew Parnell$^{2, 4}$, Dermot Brabazon$^{3, 4}$, Alessio Benavoli$^1$}}
\affil{\small{$^1$School of Computer Science and Statistics, Trinity College Dublin, Ireland\\
$^2$Hamilton Institute, Maynooth University, Ireland\\
$^3$School of Mechanical \& Manufacturing Engineering, Dublin City University, Ireland\\
$^4$I-Form Advanced Manufacturing Research Centre, Science Foundation Ireland}}

\begin{document}
\maketitle
\begin{abstract}
Bayesian optimization (BO) is an approach to globally optimizing black-box objective functions that are expensive to evaluate. BO-powered experimental design has found wide application in materials science, chemistry, experimental physics, drug development, etc. This work aims to bring attention to the benefits of applying BO in designing experiments and to provide a BO manual, covering both methodology and software, for the convenience of anyone who wants to apply or learn BO. In particular, we briefly explain the BO technique, review all the applications of BO in additive manufacturing, compare and exemplify the features of different open BO libraries, unlock new potential applications of BO to other types of data (e.g., preferential output). This article is aimed at readers with some understanding of Bayesian methods, but not necessarily with knowledge of additive manufacturing; the software performance overview and implementation instructions are instrumental for any experimental-design practitioner. Moreover, our review in the field of additive manufacturing highlights the current knowledge and technological trends of BO. This article has a supplementary material online.
\end{abstract}

\section{Introduction}
\begin{figure}[!h]
  \centering
  \includegraphics[width=15cm]{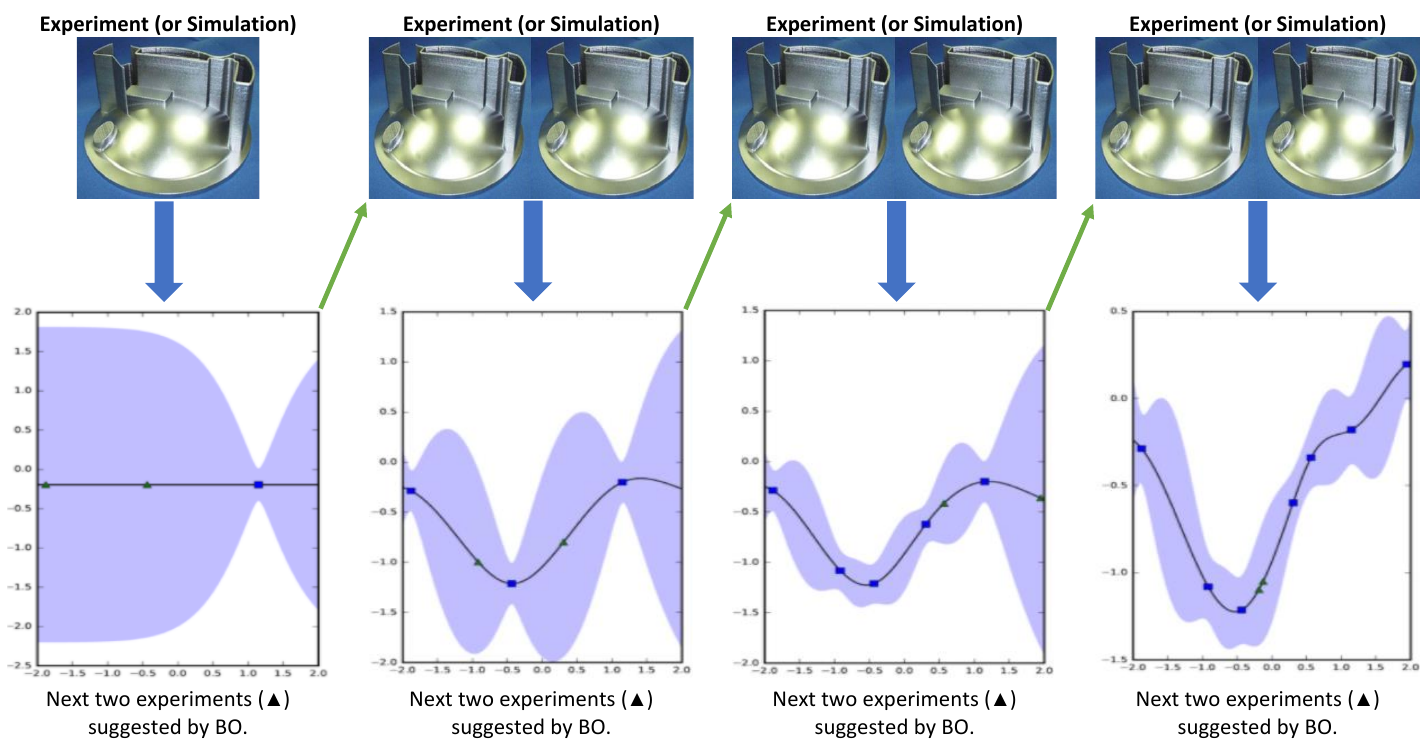}
  \caption{Given preliminary and tentative experimental data, BO will learn a model and decide the next experiment or batch of experiments. After new experimental data are available, BO will update the model and again decide the next experiment or batch of experiments. The procedure iterates until the resource budget is spent. In the above artificial AM example, we initially have one experimental output (indicated by $\blacksquare$); after learning a model (the straight line), BO suggests the next two experimental trials (indicated by $\blacktriangle$).  After the two products are printed,  BO updates the model (into a curve) according to the three available experimental outputs (indicated by $\blacksquare$ in the 2nd bottom panel), and then suggests another two experimental trials (indicated by $\blacktriangle$ in the 2nd bottom panel). The procedure thus iterates by performing suggested experiments, updating the model, and making new suggestions. The light-blue shaded bands in each bottom panel represent the prediction interval (i.e. uncertainty). Product image is from Optomec (http://www.optomec.com/).}\label{Bayesian_optimization}
\end{figure}

Engineering designs are usually performed under strict budget constraints. Collecting a single datum from computer experiments such as computational fluid dynamics can potentially take weeks or months. Each datum obtained, whether from a simulation or a physical experiment, needs to be maximally informative of the goals we are trying to accomplish. It is thus crucial to decide where and how to collect the necessary data to learn most about the subject of study. Data-driven experimental design appears in many different contexts in chemistry and physics \citep[e.g.][]{LamAerospace} where the design is an iterative process and the outcomes of previous experiments are exploited to make an informed selection of the next design to evaluate. Mathematically, it is often formulated as an optimization problem of a black-box function (that is, the input-output relation is complex and not analytically available).  Bayesian optimization (BO) is a well-established technique for black-box optimization and is primarily used in situations where (1) the objective function is complex and does not have a closed form, (2) no gradient information is available, and (3) function evaluations are expensive \citep[see][for a tutorial]{frazier2018tutorial}. BO has been shown to be sample-efficient in many domains \citep[e.g.][]{vahid2018new, Deshwalnanoporous, turner2021bayesian}. To illustrate the BO procedure, consider the (artificial) problem in additive manufacturing (AM), where we want to optimize the porosity of the printed product with respect to only one process parameter (e.g. laser power). Figure \ref{Bayesian_optimization}
demonstrates the iterative experiment-design loop, where the upper panels represent performing BO-suggested experiments, and the lower panels represent data-driven design selection.

\begin{figure}[!h]
  \centering
  \includegraphics[width=9cm]{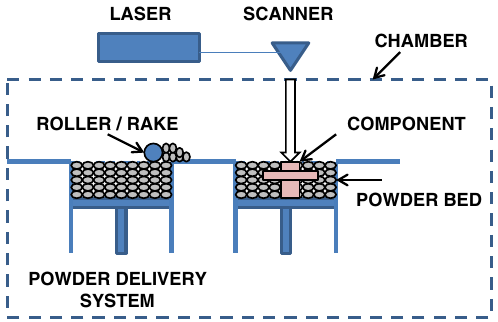}
  \caption{Generic illustration of an AM powder bed system. Reprinted from \cite{Frazier20141917}.}\label{SLM}
\end{figure}

\begin{figure}[!h]
  \centering
  \includegraphics[width=12cm]{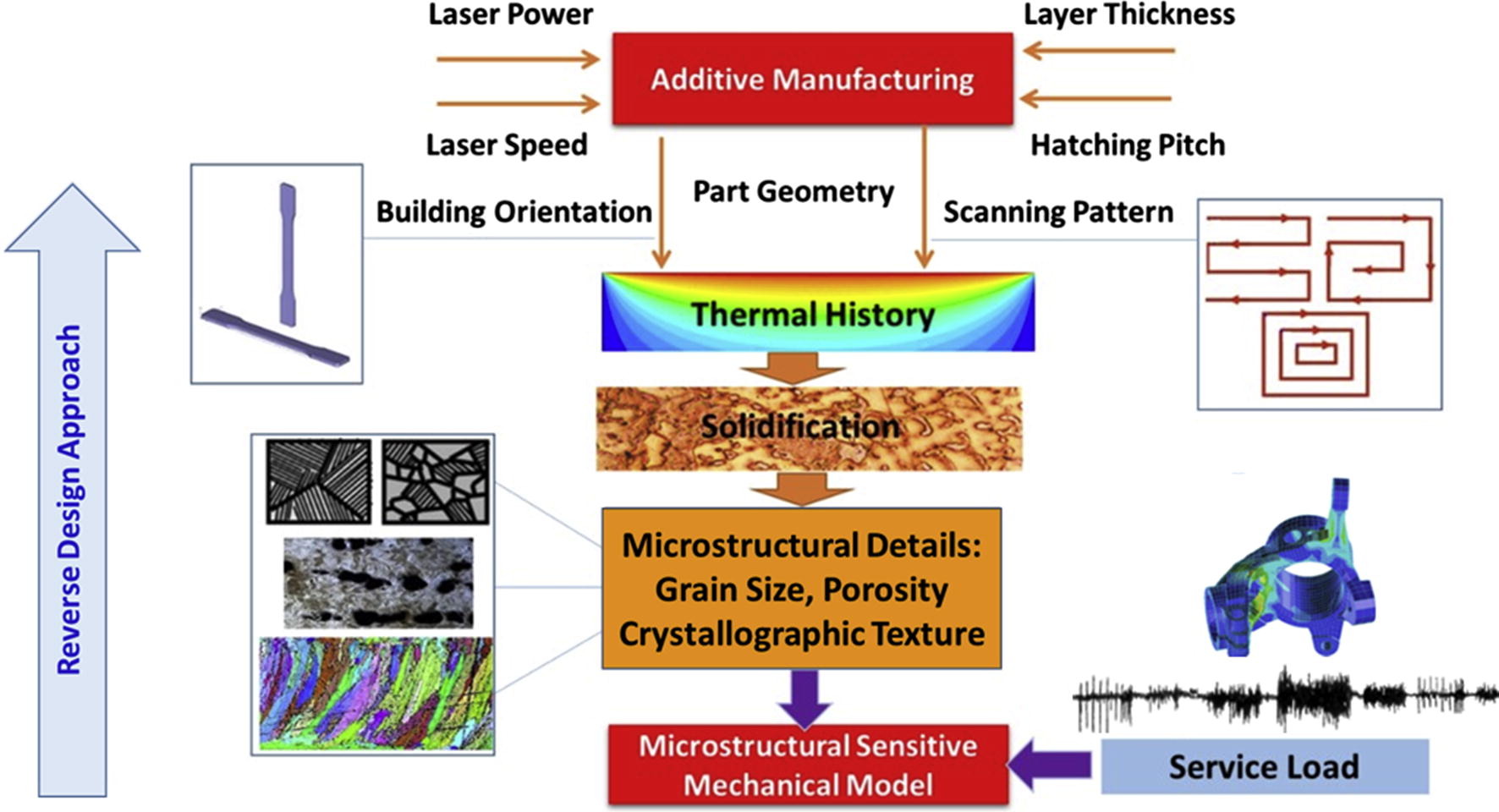}
  \caption{There are many factors (namely design parameters) that will jointly affect the structural properties of fabricated parts. Process parameters affect the thermal history of the AM parts. The thermal history during fabrication governs solidification, and consequently all the resultant microstructural details. Finally, the microstructural features dictate the structural properties of fabricated parts. Reprinted from \cite{YADOLLAHI201714}, with permission from Elsevier.}\label{flow}
\end{figure}

AM technologies have been developed in the past few years from rapid prototyping to mass scale production-ready approaches. A common AM technique: selective laser melting (aka powder bed fusion), is schematically shown in Figure \ref{SLM}.
This process involves spreading a thin layer of loose powder over a platform, and then a fibre laser or electron beam tracing out a thin 2D cross-sectional area of the part, melting and solidifying the particles together. The platform is then lowered, and another powder layer is deposited. The process is repeated as many times as needed to form the full 3D part. The main challenge against the continued adoption of AM by industries is the uncertainty in structural properties of the fabricated parts. Taking metal-based AM as an example, in addition to the variation in as-received powder characteristics, building procedure and AM systems, this challenge is exacerbated by the many involved process parameters, such as laser power, laser speed, layer thickness, etc., which affect the thermal history during fabrication. Thermal history in AM process then affects surface roughness and microstructural, and consequently mechanical behavior of fabricated parts; see Figure \ref{flow}.

The advent of AM has brought about radically new ways of manufacturing end-use products, by exploiting the space of computer-aided designs and process parameters. However, most previous AM experiments did not use BO to guide the selection of experiments, although simulations have revealed that using BO would be more efficient. Compared with domains like materials science, BO has received very little attention in the AM field. In this work, the experimental design space is a combination of product structural design (e.g. parametric design information, mesh data, material information, etc.) and manufacturing process parameters (e.g. laser power, layer thickness, hatching pitch, etc.). Finding the optimal design of an experiment can be challenging. A structured and systematic approach is required to effectively search the enlarged design space. Here we aim to bring attention to the benefits of applying BO in AM experiments and to provide a BO manual, covering both methodology and software, for the convenience of anyone (not limited to AM practitioners) who wants to apply or learn BO for experimental design. Our contributions include
\begin{itemize}
  \item A thorough review of the literature on the application of BO to AM problems.
  \item An elaborate introduction to prominent open-source software, highlighting their core features.
  \item A detailed tutorial on implementing different BO software to solve different DoE problems in AM.
  \item An illustration of novel application of BO, where the output data are preference data.
\end{itemize}

The rest of the paper is organised as follows. In Section \ref{BO-GP}, we introduce the ingredients of BO: a probabilistic surrogate model and an acquisition function. In Section \ref{AMreview}, we provide a review of the varied successful applications of BO in AM. In Section \ref{Software}, we summarize popular open-source packages implementing various forms of BO. Section \ref{Examples} provides code examples for batch optimization, multi-objective optimization and optimization with black-box constraints, and further illustrates a recently proposed extension of the basic BO framework to preference data. We summarize our findings in Section \ref{Conc}.

\section{Bayesian Optimization}\label{BO-GP}
\begin{figure}[!h]
  \centering
  \includegraphics[width=12cm]{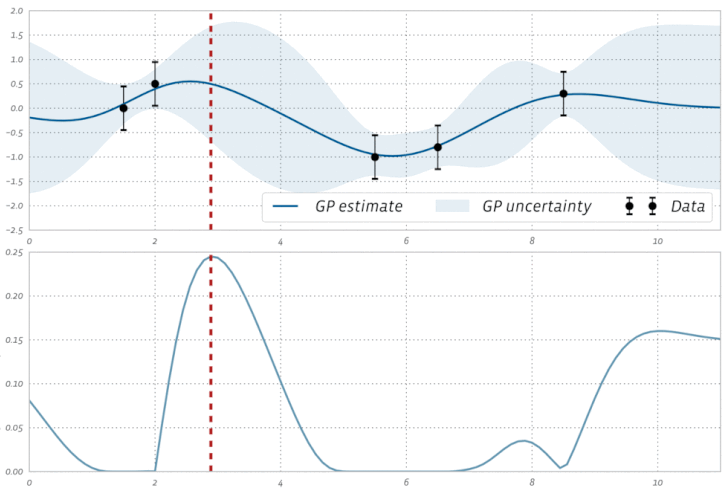}
  \caption{Upper panel: A surrogate model is fitted to five noisy observations using GPs to predict the objective (solid line) and place uncertainty estimates (proportional to the width of the shaded bands) over the range of possible parameter values. Lower panel: The predictions and the uncertainty estimates are combined to derive an acquisition function which quantifies the utility of running another experiment at a parameter value. The parameter value that maximizes the acquisition function (red dashed line) will be tested in the next experiment. The surrogate model and then the acquisition function will be updated with the results of that experiment. Figure produced by Facebook Ax.}\label{GP_AF}
\end{figure}

In this section we provide a brief introduction to BO. A more complete treatment can be found in \cite{jones1998efficient} and \cite{frazier2018tutorial}. A review of BO and its applications can be found in \cite{7352306} and \cite{8957442}. In general, BO is a powerful tool for optimizing ``black-box'' systems of the form:
\begin{equation}\label{BOp}
\mathbf{x}^*=\arg\max\limits_{\mathbf{x}\in \mathscr{X}} f(\mathbf{x}),
\end{equation}
where $\mathscr{X}$ is the design space, and the objective function $f(\mathbf{x})$ does not have a closed-form representation, does not provide function derivatives, and only allows point-wise evaluation. The value of the black-box function $f$ at any query point $\mathbf{x}\in \mathscr{X}$ can be obtained via, e.g. a physical experiment. However, the evaluation of $f$ could be corrupted by noise. Thus the output $y$ we many receive at point $\mathbf{x}$ is:
\begin{equation*}
y=f(\mathbf{x})+\epsilon,
\end{equation*}
where the noisy term $\epsilon$ is usually assumed to be normally distributed with mean 0 and variance $\sigma^2$.
A BO framework has two elements: a surrogate model for approximating the black-box function $f$, and an acquisition function for deciding which point to query for the next evaluation of $f$. Here we employ the Gaussian Process (GP) model as the surrogate due to its popularity,  but other models such as Bayesian neural networks are also commonly used as long as they provide a measure of uncertainty; see \cite{Forrester2008} for an overview. The pseudo code of the BO framework is given in Algorithm \ref{BOalg}, where we iterate between steps 2 and 5 until the evaluation budget is over.
\begin{algorithm}
\caption{Bayesian optimization}
\label{BOalg}
\begin{algorithmic}[1]
\For{$n=1, 2, 3, \ldots$,}
\State select new $\mathbf{x}_{n+1}$ by optimizing the acquisition function $\alpha$:
\begin{equation*}
  \mathbf{x}_{n+1}=\arg\max\limits_{\mathbf{x}\in \mathscr{X}}\alpha (\mathbf{x}; \mathscr{D}_{1:n});
\end{equation*}
\State query the objective function to obtain $y_{n+1}$;
\State augment data $\mathscr{D}_{1:(n+1)}=\{\mathscr{D}_{1:n}, (\mathbf{x}_{n+1}, y_{n+1})\}$;
\State update the GP model;
\EndFor
\end{algorithmic}
\end{algorithm}
A 1D example in Figure \ref{GP_AF}
further explains Algorithm \ref{BOalg}, where the upper panel corresponds to model fitting (steps 4\&5), and the lower panel corresponds to deciding the next experiment (steps 2\&3).

A GP is determined by its mean function $m(\mathbf{x})=\mbox{E}[f(\mathbf{x})]$ and covariance function $k(\mathbf{x}, \ddot{\mathbf{x}})=\mbox{E}[(f(\mathbf{x})-m(\mathbf{x}))(f(\ddot{\mathbf{x}})-m(\ddot{\mathbf{x}}))]$. For any finite collection of input points $\mathbf{x}_{1:n}=\{\mathbf{x}_1, \ldots, \mathbf{x}_n\}$, the (column) vector of function values $f(\mathbf{x}_{1:n})$ follows the normal distribution $N(m(\mathbf{x}_{1:n}), k(\mathbf{x}_{1:n}, \mathbf{x}_{1:n}))$. Here, we employ compact notation for functions applied to collections of input points: $f(\mathbf{x}_{1:n})=[f(\mathbf{x}_1), \ldots, f(\mathbf{x}_n)]^T$, $m(\mathbf{x}_{1:n})=[m(\mathbf{x}_1), \ldots, m(\mathbf{x}_n)]^T$, and $k(\mathbf{x}_{1:n}, \mathbf{x}_{1:n})$ is the $n\times n$ covariance matrix with $k(\mathbf{x}_i, \mathbf{x}_j)$ being the (i, j)th element. The covariance function $k(\mathbf{x}, \ddot{\mathbf{x}})$ controls the smoothness of the stochastic process. It has the property that points $\mathbf{x}$ and $\ddot{\mathbf{x}}$ that are close in the input space have a large correlation, encoding the belief that they should have more similar function values than points that are far apart. A popular choice for the covariance function is the class of Matern kernels. Matern kernels are parameterized by a smoothness
parameter $\nu>0$, and samples from a GP with a higher $\nu$ value are more smoother. The following are the most commonly used Matern kernels:
\begin{eqnarray*}
k_{\nu=\frac{1}{2}}(\mathbf{x}, \ddot{\mathbf{x}}) &=& \theta_0^2\exp(-r), \\
k_{\nu=\frac{3}{2}}(\mathbf{x}, \ddot{\mathbf{x}}) &=& \theta_0^2\exp(-\sqrt{3}r)(1+\sqrt{3}r), \\
k_{\nu=\frac{5}{2}}(\mathbf{x}, \ddot{\mathbf{x}}) &=& \theta_0^2\exp(-\sqrt{5}r)(1+\sqrt{5}r+\frac{5}{3}r^2),
\end{eqnarray*}
where $r=(\mathbf{x}-\ddot{\mathbf{x}})^T\Lambda(\mathbf{x}-\ddot{\mathbf{x}})$, and $\Lambda$ is a diagonal matrix of squared length scales $\theta_i^2$.

For the query points $\mathbf{x}_{1:n}$, let $y_{1:n}$ denote the vector of noisy outputs and write $\mathscr{D}_{1:n}=\{(\mathbf{x}_i, y_i): i=1, \ldots, n\}$. Conditional on the data $\mathscr{D}_{1:n}$, we now want to infer the value of $f$ at a new point $\mathbf{x}\in \mathscr{X}$. Our prior belief about the black-box function $f$ is the GP, which implies that the prior $p(f(\mathbf{x}), f(\mathbf{x}_{1:n}))$ is normal. The likelihood for the outputs $y_{1:n}$ is a normal distribution $p(y_{1:n}|f(\mathbf{x}_{1:n}))= N(y_{1:n}; f(\mathbf{x}_{1:n}), \sigma^2 I)$, where  I is the identity matrix. We can thus create the joint distribution:
\begin{equation*}
p(f(\mathbf{x}), f(\mathbf{x}_{1:n}), y_{1:n})=p(f(\mathbf{x}), f(\mathbf{x}_{1:n}))p(y_{1:n}|f(\mathbf{x}_{1:n})),
\end{equation*}
which is also a normal distribution. Therefore, the prior over outputs $y_{1:n}$ and target $f(\mathbf{x})$ is normal:
\begin{equation*}
p(f(\mathbf{x}), y_{1:n})= N\left(\left[\begin{array}{c}
   f(\mathbf{x}) \\
   y_{1:n}
 \end{array}
\right]; \left[\begin{array}{c}
                           m(\mathbf{x}) \\
                           m(\mathbf{x}_{1:n})
                         \end{array}
\right], \left[\begin{array}{cc}
                 k(\mathbf{x}, \mathbf{x}) & k(\mathbf{x}, \mathbf{x}_{1:n}) \\
                 k(\mathbf{x}_{1:n}, \mathbf{x}) & k(\mathbf{x}_{1:n}, \mathbf{x}_{1:n})+\sigma^2 I
               \end{array}
\right]\right).
\end{equation*}
Utilizing the rule for conditionals, the distribution $p(f(\mathbf{x})|\mathscr{D}_{1:n}, \mathbf{x})$ is normal, with mean
\begin{equation*}
m_n(\mathbf{x})=m(\mathbf{x})+k(\mathbf{x}, \mathbf{x}_{1:n})[k(\mathbf{x}_{1:n}, \mathbf{x}_{1:n})+\sigma^2 I]^{-1}(y_{1:n}-m(\mathbf{x}_{1:n})),
\end{equation*}
and variance
\begin{equation*}
\sigma_n^2(\mathbf{x})=k(\mathbf{x}, \mathbf{x})-k(\mathbf{x}, \mathbf{x}_{1:n})[k(\mathbf{x}_{1:n}, \mathbf{x}_{1:n})+\sigma^2 I]^{-1}k(\mathbf{x}_{1:n}, \mathbf{x}).
\end{equation*}
That is, after observing the data, our posterior belief about the black-box function $f$ is still a GP.

The acquisition function is typically a computationally inexpensive function, evaluating how desirable a query point $\mathbf{x}$ is for the optimization problem (\ref{BOp}). In other words, we replace our original optimization problem with another optimization problem, but on a cheaper function. One commonly used acquisition function is called expected improvement (EI). By maximizing the EI acquisition function, the next query point at which to evaluate $f$ is the one that, in expectation, improves upon $f$ the most. Let $\hat{f}=\max_{\mathbf{x}_i\in\mathbf{x}_{1:n}} f(\mathbf{x}_i)$ be the maximal value of $f$ observed so far. For any query point $\mathbf{x}$, our utility would be $\max(0, f(\mathbf{x})-\hat{f})$; that is, we receive a reward equal to the improvement $f(\mathbf{x})-\hat{f}$ if $f(\mathbf{x})$ turns out to be larger than $\hat{f}$, and no reward otherwise. The EI acquisition function is then the expected utility at $\mathbf{x}$:
\begin{align*}
\alpha (\mathbf{x}; \mathscr{D}_{1:n})&=\mbox{E}[\max(0, f(\mathbf{x})-\hat{f})|\mathscr{D}_{1:n}]\\
&=\int_{\hat{f}}^{\infty} (f-\hat{f})p(f|\mathscr{D}_{1:n}, \mathbf{x})df\\
&=\int_{\hat{f}}^{\infty} (f-\hat{f})\phi\left(\frac{f-m_n(\mathbf{x})}{\sigma_n(\mathbf{x})}\right)df\\
&=[m_n(\mathbf{x})-\hat{f}]\Phi\left(\frac{\hat{f}-m_n(\mathbf{x})}{\sigma_n(\mathbf{x})}\right)+k(\mathbf{x}, \mathbf{x})\phi\left(\frac{\hat{f}-m_n(\mathbf{x})}{\sigma_n(\mathbf{x})}\right),
\end{align*}
where $\Phi$ and $\phi$ are the CDF and PDF of the standard normal distribution, respectively. The EI acquisition function $\alpha (\mathbf{x}; \mathscr{D}_{1:n})$ has two components. The first can be increased by increasing the mean function $m_n(\mathbf{x})$. The second can be increased by increasing the variance $k(\mathbf{x}, \mathbf{x})$. These two terms can be interpreted as encoding a trade-off between exploitation (evaluating at points with high mean) and exploration (evaluating at points with high uncertainty).

Another three popular acquisition functions are entropy search (ES), upper confidence bound (UCB) and knowledge gradient (KG). ES attempts to maximize our information about the location of the maximum $\mathbf{x}^*$ given new data.  UCB is often described in terms of maximizing $f$ rather than minimizing $f$. In the context of minimization, the acquisition function is called lower confidence bound, but UCB is ingrained in the literature as a standard term. The UCB acquisition function can be expressed as $\alpha (\mathbf{x}; \mathscr{D}_{1:n})=m_n(\mathbf{x})+\beta \sigma_n(\mathbf{x})$, where $\beta$ is a tuning parameter. The query point maximizing the UCB corresponds to a (fixed-probability) best case scenario. The KG criterion quantifies the expected increase in the maximum of the surrogate model from obtaining new data. Recently, a flurry of acquisition functions for batch optimization have been developed. These include Thompson sampling, local penalization, and extensions of classical acquisition functions such as qKG, dKG, qEI, bUCB, etc. A short survey of acquisition functions is given by \cite{NathanR}.\footnote{A more complete list of acquisition functions can be found in the webpage ``Acquisition Function APIs'' of BoTorch (https://botorch.org/api/acquisition.html).} An acquisition function decides how to explore the parameter space, and there is no one-size-fits-all acquisition function. The choice of acquisition functions depends on the problem at hand and the trade-off between exploitation and exploration: exploitation focuses on results in the vicinity of the current best results, while exploration pushes the search towards unexplored regions.

\section{Review of Bayesian Optimization in Additive Manufacturing}\label{AMreview}
We performed a systematic search of the literature for the use of BO in AM, using the keywords ``Bayesian'', ``sequential'', ``adaptive'', ``additive manufacturing'' and ``3D printing''. Search for the literature was carried out via Scopus and Google Scholar. Research on applying Bayesian optimization to additive manufacturing is still at an early stage with limited literature. Only eight papers were found, and all of them were published in the last five years.

Mechanical metamaterials are artificial structures defined by their geometry rather than their chemical composition. One class of metamaterials is structural lattices. Three previous papers have applied BO for lattice structure design. To reduce the number of design variables, \cite{DETC2018-85270} utilized the geometric projection technique and represented the struts in a lattice structure by rod primitives. The authors developed a constrained BO framework for micro-architecture (aka, unit cell) design, to achieve a high stiffness-to-weight ratio. Computationally expensive constraints are approximated by surrogate models, and the probabilities of satisfying the expensive constraints are incorporated as weights into the expected improvement function. Their study used two rods of equal diameter, yielding 13 total design variables. After the optimization process, the optimal design is obtained by collating the endpoints of bars exactly and extending the bars to the boundaries of the unit cell domain where necessary. The BO framework successfully identified a number of interesting unit cell designs that compare favorably with the popular octet truss design.
\cite{IMECE2020-23377} studied both unit cell design optimization (with design variables: height, width, and strut thickness) and lattice structure design optimization (with two more design variables: number of rows and columns of unit cells). Both optimization problems have the same objective function: a weighted sum of the head injury criterion and part volume. The work focuses on two types of unit cells (diamond and honeycomb) which are parameterized by geometric attributes including height, width, and strut thickness. Due to the computational expense of nonlinear finite element analysis and the difficulty of obtaining objective gradient information, the GP model is applied to approximate the objective function. The GP model is initialized with four random training points in the unit cell design optimization, and twelve random points in the lattice structure design optimization.
Utilizing multi-material 3D printing techniques (designing novel composite solids by combining multiple base materials with distinct properties), it is possible to design mechanical metamaterials by varying spatial distributions of different base materials within a representative volume element (RVE), which is then periodically arranged into a lattice structure. In \cite{XUE2020100992}, the goal is to design an RVE composed of different base materials so that when the RVEs are periodically arranged into a lattice structure, the macroscopic elastic moduli achieve a desired set of values. They adopted variational autoencoders to compress RVE images that describe the spatial distribution of base materials to a reduced latent space, and then performed BO over the reduced space to find an optimal RVE configuration that fulfills the design goal. The objective function measures the distance to the targeted Young's modulus and Poisson's ratio. The computational cost to evaluate the objective function is not cheap, so they apply a noise-free GP as the surrogate and use expected improvement as the acquisition function.

\cite{Pahl2007} defined a design process with four steps: requirement clarification, conceptual design, embodiment design, and detailed design. In embodiment design (aka, design exploration), designers determine the overall layout design, the preliminary form design, and the production processes based on design feasibility checks. In detailed design (aka, design exploitation), designers narrow down a set of potential solutions to one optimal design and provide ready-to-execute instructions. \cite{1.4043587} focused on the embodiment design and detailed design steps in the context of AM. They applied the Bayesian network classifier model for embodiment design, and the GP model for detailed design. The first batch of samples, collected by the Latin hypercube sampling method, is utilized in both the embodiment design and the detailed design steps to train the two machine learning models. A computationally-expensive high-fidelity model is used to generate the corresponding responses. In the detailed design step, after training the GP model on the first batch of samples, they applied a Markov chain Monte Carlo (MCMC) sampling method to generate the second batch of design samples, evaluated the design samples via the high-fidelity model, and finally updated the GP model. Informed by the updated GP model, an optimal design is identified to meet certain targets. The method was exemplified by designing an ankle brace that includes tailored horseshoe structures in different regions to achieve stiffness requirements in the rehabilitation process. The initial batch contains 200 design samples distributed in a five-dimensional space, and the MCMC sampling method generates another batch of 35 design samples. We note that the whole design process in \cite{1.4043587} is not sequential, but the detailed design can be readily modified to be adaptive.

During layer-by-layer fabrication, with the thermal gradient and the solidification velocity in the melt pool changing continuously, the melt pool geometry undergoes substantial changes if the process parameters are not appropriately adjusted in real time. In \cite{met10050683}, the goal is to dynamically determine the values of process parameters (laser power and linear scan velocity) to minimize the deviation from the target melt pool dimensions (melt pool depth and width). As the true functional form that relates the process parameters to the melt pool dimensions is unknown, GP is applied as the surrogate model. At each control time point, the Latin hypercube sampling method is used to generate 20 training samples, and Eagar-Tsai's model is used to generate the responses; the subsequent samples are determined by the expected improvement criterion; among the evaluated samples, the one with the optimal objective value provides the values of the process parameters. At the next time point of control, the whole process repeats.

Both \cite{Gongoraeaaz1708} and \cite{Deneault2021} attempted to build an autonomous closed-loop AM system to accelerate the process of learning optimal printing conditions. The system developed by \cite{Gongoraeaaz1708} consists of five dual extruder fused-deposition-modeling printers, a scale (CP225D, Sartorius) and a universal testing machine (5965, Instron Inc.). The goal is to design the best crossed barrel in terms of toughness. A crossed barrel has two platforms that are held apart by $n$ hollow columns of outer radius $r$ and thickness $t$ and that are twisted with an angle $\theta$; hence, the design space is four-dimension. \cite{Gongoraeaaz1708} first performed a grid-based experimental campaign in which 600 distinct designs were tested in triplicate (without human intervention). Using the experimental data, they then ran a series of simulations and found that BO can reduce the number of experiments needed by a factor of almost 60. 
\cite{Deneault2021} built another autonomous system with material extrusion-type printers, which utilizes in-line image capture and analysis for feedback and BO for adaptive design. The design variables are four syringe extrusion parameters (i.e. prime delay, print speed, x-position, and y-position), and the goal is to print lines with the leading segment most closely matching the target geometry. The image of the leading segment of a line is compared to the target geometry, and an ``objective score'' is returned as the response. The prototype achieved near-ideal production of the user-defined geometry within 100 experiments.

\section{Review of Bayesian Optimization Software}\label{Software}
As of this writing, there are a variety of Bayesian optimization libraries out there. We here compare certain prominent packages that are available in R or Python.\footnote{There are also Matlab packages like ``bayesopt'', but in this review we have only focused on open software.}
R packages include:
\begin{itemize}
  \item DiceOptim is a popular R package for sequential or batch optimization. The function \emph{max\_qEI} is for batch optimization, \emph{noisy.optimizer} for noisy-output problems, and \emph{EGO.cst} for constrained problems.
  \item laGP is mainly for GP modelling and inference. It provides a wrapper routine, i.e. \emph{optim.auglag}, for optimizing black-box functions under under multiple black-box constraints.
  \item mlrMBO allows user-defined acquisition functions, generated with the \emph{makeMBOInfillCrit} function. The functions \emph{initSMBO} and \emph{updateSMBO} are for designing physical experiments (i.e. step-by-step optimization with external objective evaluation).
\end{itemize}
Python packages include:
\begin{itemize}
  \item Spearmint (https://github.com/HIPS/Spearmint) is one of the earliest user-friendly BO libraries. It is able to deal with constrained problems (the PESC branch) and multi-objective problems (the PESM branch). However, the latest version comes without the functionality for step-by-step optimization with external objective evaluation, and hence is only suitable for problems with callable objective functions.
  \item GPyOpt (https://github.com/SheffieldML/GPyOpt) is a popular Python library for optimizing physical experiments (sequentially or in batches). When optimizing the acquisition function, we can fix the value of certain variables. These variables are called context as they are part of the objective but are fixed when the acquisition function is optimized. GPyOpt allows to incorporate function evaluation costs (e.g. time) in the optimization.
  \item Cornell-MOE (https://github.com/wujian16/Cornell-MOE) is built on the MOE package. It is written in Python, with internal routines in C++. It supports box and simplex constraints. Although Cornell-MOE is easier to install (compared with the MOE package), it is only available for the Linux platform. While the MOE package allows optimization with external objective evaluation (via the function \emph{gp\_next\_points}), the Cornell-MOE package only works on callable functions. Cornell-MOE supports continuous-fidelity optimization via knowledge gradient \cite{wu2018continuous}.
  \item GPflowOpt (https://github.com/GPflow/GPflowOpt) is built on the Python library GPflow (pre-2020 versions), a package for running GP modeling tasks on a GPU using Tensorflow. This makes GPflowOpt an ideal optimizer if GPU computational resources are available. It supports implementing user-defined acquisition functions.
  \item pyGPGO (https://github.com/josejimenezluna/pyGPGO) supports integrated acquisition functions and is able to optimize covariance function hyperparameters in the GP model. More information on this package can be found from one contributor's master thesis \cite{JimenezLuna}.
  \item Emukit (https://github.com/amzn/emukit) provides a way of interfacing third-part modelling libraries (e.g. a wrapper for using a model created with GPy). When new experimental data are available, it can decide whether hyper-parameters of the surrogate model need updating based on some internal logic.
  \item Dragonfly (https://github.com/dragonfly/dragonfly) library provides an array of tools to scale up BO to expensive large scale problems. It allows specifying a time budget for optimisation. The ask-tell interface in Dragonfly enables step-by-step optimization with external objective evaluation.
  \item Trieste (https://github.com/secondmind-labs/trieste) is built on the Python library GPflow (2.x version), serving as a new version of GPflowOpt. As of this writing, it is not recommended to install both Trieste and any other BO library that depends on GPy (e.g. GPyOpt); otherwise, we will get numpy-incompatible errors. The library is under active development, and the functions may change overtime.
  \item BoTorch (https://github.com/pytorch/botorch) and Ax are two BO tools developed by Facebook. BoTorch supports both analytic and (quasi-) Monte-Carlo based acquisition functions. It provides an interface for implementing user-defined surrogate models, acquisition functions, and/or optimization algorithms. BoTorch has been receiving increasing attention from BO researchers and practitioners. Compared with BoTorch, Ax is relatively easier to use and targets at end-users.
\end{itemize}

\begin{figure}[!h]
  \hspace{-1cm}
  \includegraphics[width=19cm]{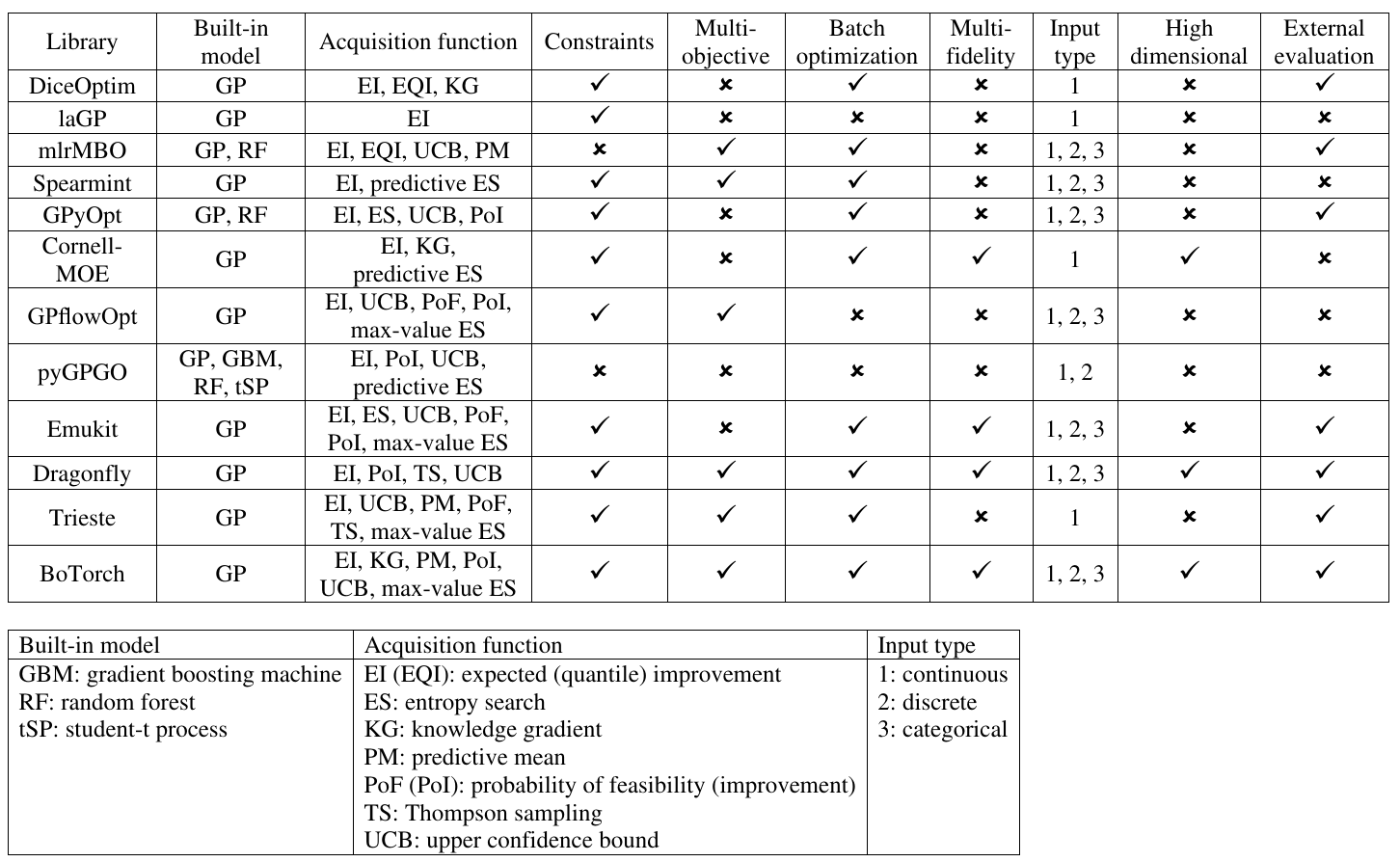}
  \caption{Comparing different libraries with respect to if they can deal with non-box constrained problems, multi-objective problems, or high dimensional problems, and if they support batch optimization, multi-fidelity optimization, categorical input, or step-by-step optimization with external objective evaluation.}\label{PKGcompare}
\end{figure}

In Figure \ref{PKGcompare}
we compare the above libraries with respect to nine important features (available as of October/2021). By a ``built-in model'', we mean a wrapper function is already available for conveniently applying a surrogate model, e.g. the GPy GP wrapper in Emukit. For certain Python libraries (e.g. BoTorch), the acquisition function list is not complete; in particular, most acquisition functions for batch optimization are not included. All libraries can deal with box constraints. Multi-objective problems involve more than one objective function to be optimized simultaneously, and optimal decisions need to be taken in the presence of trade-offs between conflicting objectives. Batch optimization is not to be confused with parallel optimization/computing, where the former recommends multiple experiments at a time. Multi-fidelity optimization algorithms leverage both low- and high-fidelity data in order to speed up the optimisation process.

With many options at hand, a natural question is then which software is the most effective. Well, there is no clear winner. \cite{Kandasamy2020} compared Dragonfly, Spearmint and GPyOpt on a series of synthetic functions. On Euclidean domains, Spearmint and Dragonfly perform well across the lower dimensional tasks, but Spearmint is prohibitively expensive in high dimensions. On the higher dimensional tasks, Dragonfly is the most competitive. On non-Euclidean domains, GPyOpt and Dragonfly perform very well on some problems, but also perform poorly on others. \cite{Balandat2020} compared BoTorch, MOE, GPyOpt and Dragonfly through the problem of batch optimization (batch size = 4), on four noisy synthetic functions. The results suggest the ranking of BoTorch (OKG), GPyOpt (LP-EI), MOE (KG) and Dragonfly (GP Bandit), where we include the acquisition function in the parentheses. \cite{vinod2020constrained} showed that Emukit is efficient for solving constrained problems, where both the objective and constraint functions are unknown and are accessible only via first-order oracles. \cite{li2021openbox} compared BoTorch, GPflowOpt and Spearmint by tuning hyper-parameters of 25 machine learning models, and the average ranking is BoTorch, Spearmint and GPflowOpt. Note that this comparison does not necessarily show that one package is better than another; it rather compares the performance of their default methods.

\section{Case Studies}\label{Examples}
Practitioners have often avoided implementing BO for DoE, mainly because the lack of tutorials on selecting and implementing available software. After comparing BO libraries in Section \ref{Software}, we here provide detailed coding examples.

\subsection{Batch Optimization with External Objective Evaluation}
\subsubsection{Background}
We exemplify batch optimization via the experimental data provided by \cite{Shrestha2017491}. The study investigated the impact of four process parameters on the binder jetting AM process of 316L stainless steel, and the objective was to maximize the transverse rupture strength (Mpa) of sintered AM metal parts. Standard ASTM B528-99 samples of nominal dimensions 31.7 mm $\times$ 12.7 mm $\times$ 6.35 mm were printed using 316L stainless steel powder. The samples were cured at 190$^\circ$C for 4 hours and then sintered in vacuum. All the samples were built along the direction that is perpendicular to the direction of loading. To measure the transverse rupture strength of the samples, a 76.2 mm hardened rod was used for testing at a loading rate of 2.5 mm/min until complete rupture occurred. Four processing parameters were investigated at three levels. The processing parameters and their levels are given in Table \ref{DoE_batch}.
\begin{table}[!h]
  \centering
  \caption{Process parameters and their levels.}\label{DoE_batch}
  \begin{tabular}{llrrrrr}
    \hline
    Process parameter& Units& L1& L2& L3 &Lower & Upper\\
    \hline
    Saturation& \%& 35& 70& 100 &35 &100 \\
    Layer thickness& $\mu m$& 80& 100& 120 &80 &120 \\
    Roll speed& mm/sec& 6& 10& 14 &6 &14 \\
    Feed-to-powder ratio& & 1& 2& 3 &1 &3 \\
    \hline
  \end{tabular}
\end{table}
Taguchi L27 orthogonal array was used to design the experiments. Four samples were printed per experiment, and the response value is the average of the four transverse rupture strength values. Data can be found in their online supplementary material.

\subsubsection{Code Examples}
We aggregate the data into a file named ``BatchObj.csv'', with column names ``Saturation'', ``Layer\_thickness'', ``Roll\_speed'', ``Feed\_powder\_ratio'', and ``y'' (for transverse rupture strength). Given the 27 experimental settings and the mean response values, we want to determine the next batch of experiments, expecting that, after a few more additional experiments, we will find an optimal experimental setting such that the printed parts are of desired strength levels. Without loss of generality, we set the batch size to be two. To determine the next batch of experimental settings, we exemplify via the following 7 libraries: \{DiceOptim, mlrMBO, GPyOpt, Emukit, Dragonfly, Trieste, BoTorch\}, as they allow both batch optimization and external objective evaluation. Note that in all the code examples, we have standardized the output data before the optimization. This is mainly because certain libraries assume that the mean function of the GP model is zero.

\begin{figure}[!h]
  \centering
  \includegraphics[scale=0.85]{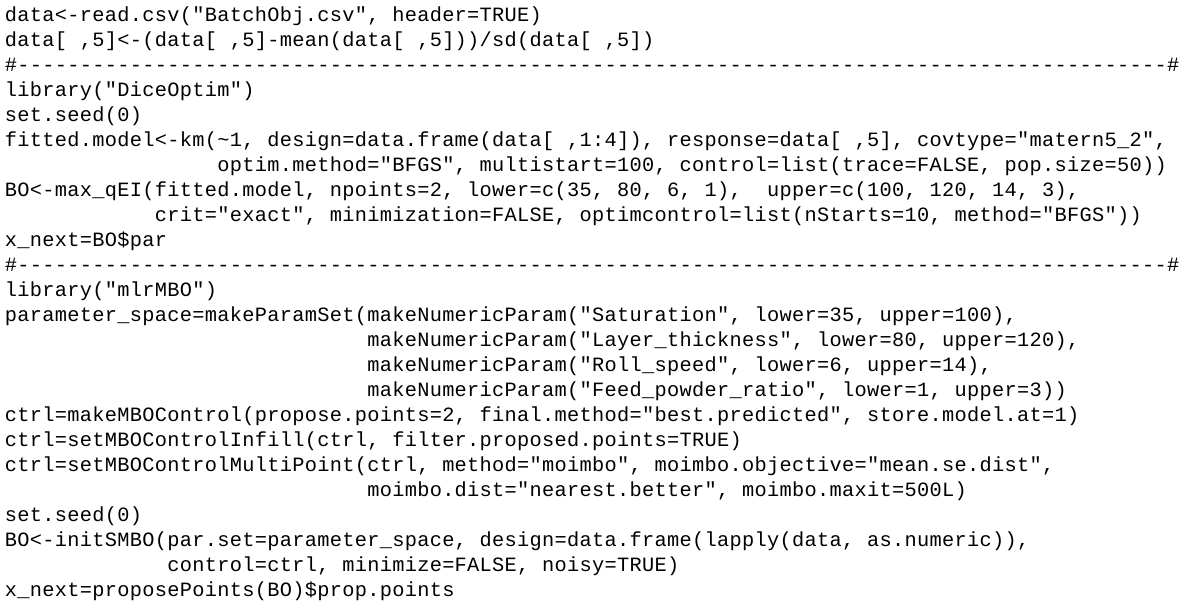}
  \caption{Code example for batch optimization using DiceOptim and mlrMBO.}\label{R_batch}
\end{figure}

\textbf{DiceOptim \& mlrMBO}: The code for batch optimization with external evaluation is given in Figure \ref{R_batch}.
For the DiceOptim package, we need to define an object of class `km' for fitting GP models. The argument `crit' in the function \textit{max\_qEI} is for specifying the method for maximizing the qEI criterion. For the mlrMBO package, we need to define an mbo control object, here denoted by `ctrl'. To define an mbo control object, we need to apply the \textit{makeMBOControl} function, and then the \textit{setMBOControlMultiPoint} function for batch optimization. Here we select `moimbo' for proposal of multiple infill points. Note that if `moimbo' is selected, the infill criterion in \textit{setMBOControlInfill} is ignored, and hence we didn't specify any value for the argument `crit' in the \textit{setMBOControlInfill} function. The method `moimbo' proposes points by multi-objective infill criteria via evolutionary multi-objective optimization. The other two options are `cb' that proposes points by optimizing the confidence bound criterion, and `cl' that proposes points by the constant liar strategy. The functionalities of DiceOptim and mlrMBO can be found from their reference manuals.

\begin{figure}[!h]
  \centering
  \includegraphics[scale=0.85]{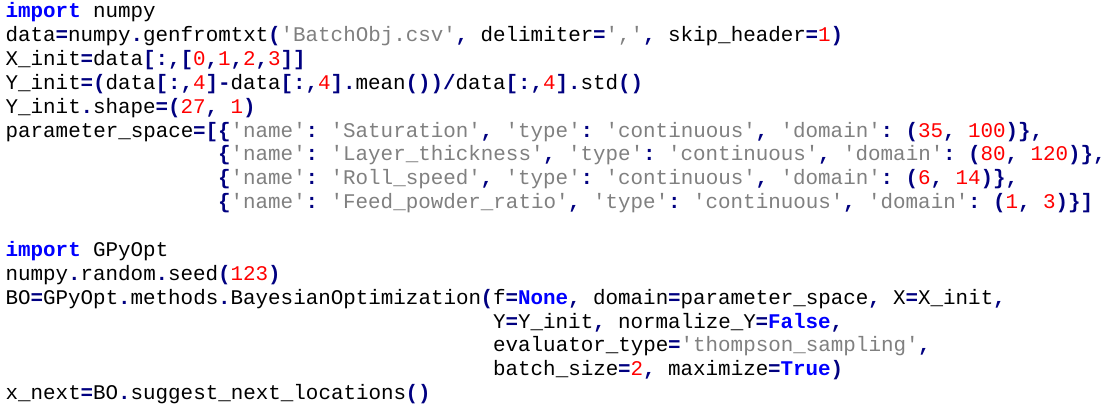}
  \caption{Code example for batch optimization using GPyOpt.}\label{GPyOpt_batch}
\end{figure}

\textbf{GPyOpt}: Compared to other Python libraries, GPyOpt is relatively simple to use, and its functionalities are well documented. The code example in Figure \ref{GPyOpt_batch}
applies Thompson sampling for batch optimization, and another option is local penalization. The argument `maximize' allows us to maximize the objective function.

\begin{figure}[!h]
  \centering
  \includegraphics[scale=0.85]{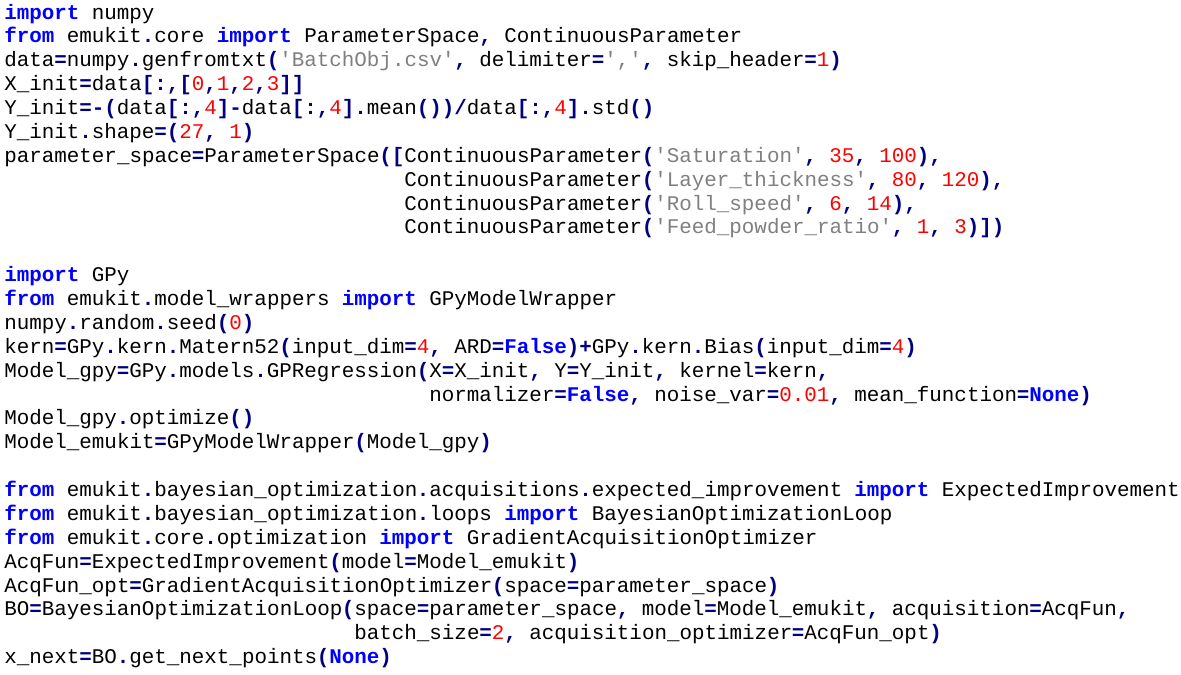}
  \caption{Code example for batch optimization using Emukit.}\label{Emukit_batch}
\end{figure}

\textbf{Emukit}: Though Emukit implements the experimental-design feature with the function \textit{ExperimentalDesignLoop}, the function \textit{BayesianOptimizationLoop} from the Bayesian-optimization feature is also suitable for step-by-step optimization with external objective evaluation. Figure \ref{Emukit_batch}
exemplifies the code. Here we set the GP kernel to be a combination of Matern-5/2 kernel and bias kernel and apply the available wrapper \textit{GPyModelWrapper} to plug GPy models into Emukit. Models written in GPy, TensorFlow, MXnet, etc. all can be wrapped in Emukit. To be consistent with the configurations of other BO libraries, we choose `expected improvement' for batch optimization; another popular choice is `local penalization'. We optimize the acquisition function using a quasi-Newton method (L-BFGS). Note that, for the GP kernel, we set the automatic relevance determination feature to be false (ARD=False). This is only to avoid the LinAlg-error of ``not positive definite'' when dealing with the ``BatchObj'' data. It is recommended to always set `ARD' to be true in practice.

\begin{figure}[!h]
  \centering
  \includegraphics[scale=0.85]{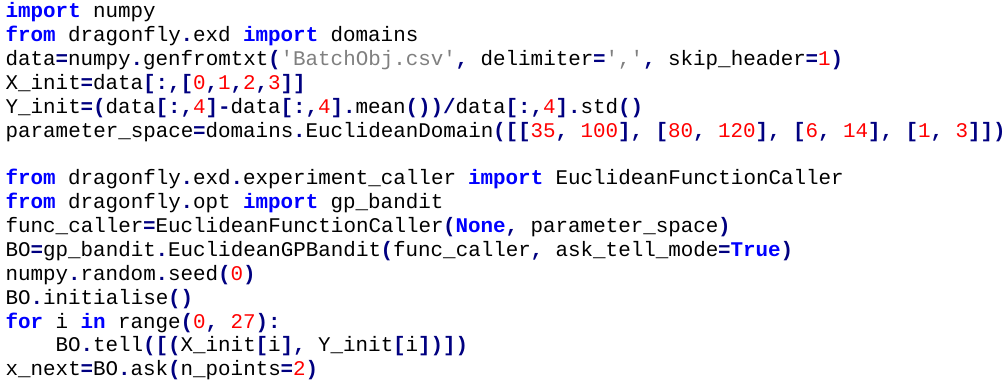}
  \caption{Code example for batch optimization using Dragonfly.}\label{Dragonfly_batch}
\end{figure}

\textbf{Dragonfly}: By default, Dragonfly maximizes functions. For a minimization problem, simply take the negative of the objective/outputs and perform the same procedure as below. The ask-tell interface in Dragonfly enables step-by-step optimization with external objective evaluation. Two main components are required in the ask-tell interface: a function caller and an optimizer. Our process parameters are continuous, and hence we adopt the \textit{EuclideanFunctionCaller} and the \textit{EuclideanGPBandit} optimizer. The code example is given in Figure \ref{Dragonfly_batch}.
Note that, even though the \textit{EuclideanGPBandit} optimizer includes the `options' argument (that allows you to select an acquisition function from \{TS, UCB, EI, top-two EI, batch UCB\}), it is ignored if the `ask\_tell\_mode' argument is set to be true. By setting `n\_points = 2', we will obtain the next two experimental settings.

\begin{figure}[!h]
  \centering
  \includegraphics[scale=0.85]{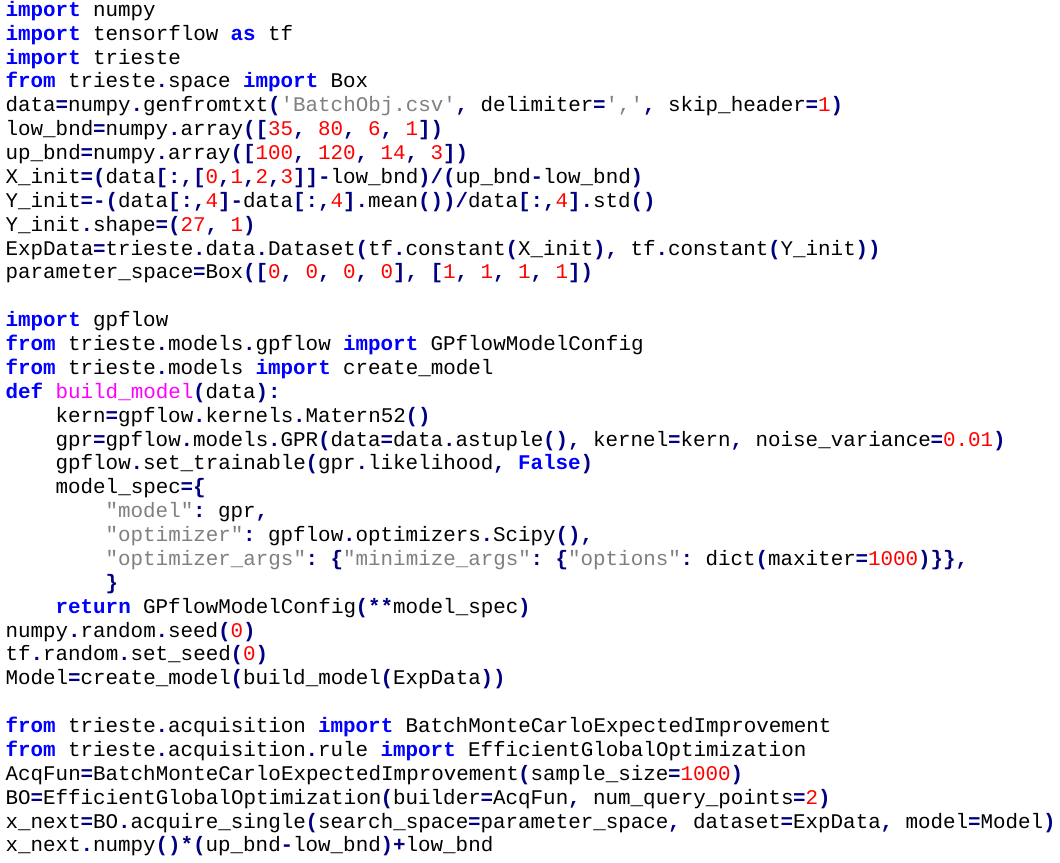}
  \caption{Code example for batch optimization using Trieste.}\label{Trieste_batch}
\end{figure}

\textbf{Trieste}: Trieste only supports minimizing the objective function, and hence we flip the sign of the output data from positive to negative. The code example is given in Figure \ref{Trieste_batch}.
Note that we have normalized the inputs by linearly mapping the input space into the unit hypercube. This is for better estimating the length-scale parameters in the Matern kernel. The function \textit{build\_model} is to configure the GP model from GPflow, and the function \textit{create\_model} returns a trainable probabilistic model built according to the configuration. Note that we need to set the likelihood to be non-trainable. Three batch acquisition functions are supported in Trieste: \textit{BatchMonteCarloExpectedImprovement}, \textit{LocalPenalizationAcquisitionFunction} and \textit{GIBBON}. \textit{BatchMonteCarloExpectedImprovement} jointly allocates the batch of points as those with the largest expected improvement over our current best solution. In contrast, the \textit{LocalPenalizationAcquisitionFunction} greedily builds the batch, sequentially adding the maximizers of the standard (non-batch) EI criterion. \textit{GIBBON} also builds batches in a greedy manner but seeks batches that provide a large reduction in our uncertainty around the maximum value of the objective function. The last line of code projects the query points in the unit hypercube back to the original input space.

\begin{figure}[!h]
  \centering
  \includegraphics[scale=0.85]{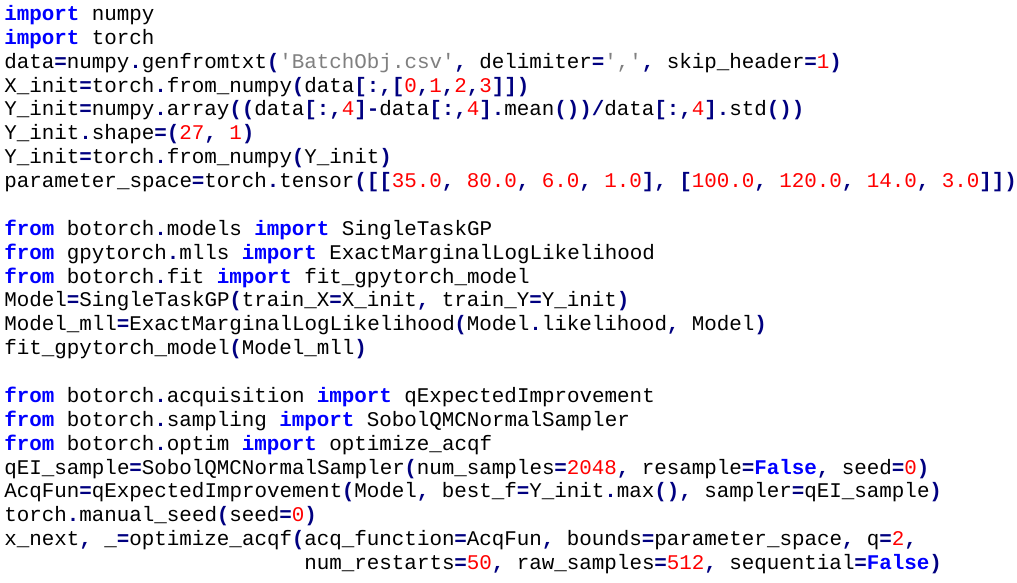}
  \caption{Code example for batch optimization using BoTorch.}\label{BoTorch_batch}
\end{figure}

\textbf{BoTorch}: BoTorch implements MC acquisition functions that support for batch optimization. They are referred to as q-acquisition functions (e.g. q-EI, q-UCB, and a few others). The code example in Figure \ref{BoTorch_batch}
applies q-EI, the MC counterpart of the EI criterion. Note that if we apply the EI criterion, then the optimization will only suggest one experimental setting. The function \textit{ExactMarginalLogLikelihood} is to compute the marginal log-likelihood of the GP model, and it works only when the likelihood is Gaussian.

We have set the seed in each code example, so that our results (given in Table \ref{BatchResults}) are reproducible.
\begin{table}[!h]
  \centering
  \caption{Batch optimization results returned by different libraries.}\label{BatchResults}
    \begin{tabular}{lcc}
    \hline
    Library& 1st query point& 2nd query point\\
    \hline
    DiceOptim& (48.51, ~95.27, 11.45, 2.46)&  (46.03, 115.69, ~9.70, 2.09)\\
    mlrMBO&    (67.55, 111.70, ~6.00, 2.97)&  (70.57, 117.53, ~6.00, 2.99)\\
    GPyOpt&    (81.55, ~85.05, 14.00, 1.00)&  (35.00, ~82.23, ~6.96, 1.14)\\
    Emukit&    (83.46, ~80.24, ~9.97, 1.98)&  (48.49, 119.38, 13.11, 1.56)\\
    Dragonfly& (93.20, ~86.49, 11.56, 2.35)&  (84.86, 114.69, ~8.63, 2.74)\\
    Trieste&   (66.63, 109.30, ~6.00, 3.00)&  (45.32, ~86.60, ~9.40, 2.36)\\
    BoTorch&   (36.28, 103.58, 12.87, 2.71)&  (70.50, 119.86, ~6.08, 1.03)\\
    \hline
  \end{tabular}
\end{table}
From Table \ref{BatchResults} we cannot detect any strange result, except that the two query points returned by mlrMBO are close to each other. The main takeaways are: (1) We would not recommend the ask-tell interface in Dragonfly, as it is of very limited flexibility. (2) BoTorch is ideal for BO practitioners, yet may require further knowledge of underlying theory/statistics for engineers. (3) For the two R libraries, DiceOptim is relatively simpler to use than mlrMBO.

\begin{figure}[!h]
  \centering
  \includegraphics[scale=0.85]{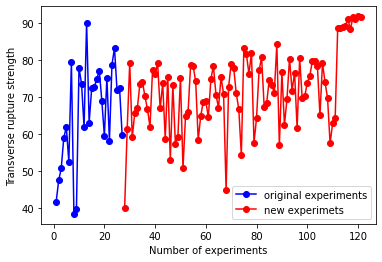}
  \caption{Evolution of the experimental results when applying the BO algorithm.}\label{trace}
\end{figure}

To help understand how and when BO gets to the optimal experimental design, we here plot the evolution of the response variable if the experiment were to continue. For illustrative purpose, we arbitrarily assume that the relationship between the response (transverse rupture strength) and the four factors (saturation, layer thickness, roll speed and feed-to-powder ratio) is multivariate polynomial with degree 2, and the unknown model parameters are estimated from the whole 108 experimental results. The BoTorch library is utilized for optimization. Once the BO algorithm produces a batch of two experimental settings, the query points are fed into the estimated polynomial model to get the outputs. The experimental data are then updated to include the new experimental results, and the updated data are fed into the BO algorithm to get the next batch of two experimental settings.  Figure \ref{trace}
plots the evolution of the response variable. The response values show a large variation at the beginning (exploration) and later cluster at a high value (exploitation). This implies that early experimental settings are spread over the whole parameter space and later experimental settings cluster around the optimum.

\subsection{Multi-objective Optimization with External Objective Evaluation}
\subsubsection{Background}
Paste extrusion is an AM technique that extrudes and deposits paste through a nozzle onto a substrate below the freezing temperature of the paste. This technique can be used to fabricate piezoelectric ceramics with better shape retention and high density. The study of \cite{Renteria_2019} aims to optimize the fabrication process of BaTiO$_3$ ferroelectric ceramics over four factors: binder amount, particle size, nozzle diameter and printing speed. The relative density, piezoelectric coefficient, dielectric permittivity and dimensional accuracy were selected as the key performance indices for the ferroelectric ceramic fabrication process. In other words, the optimization problem has four objectives. A two-level fractional factorial design with three replications was implemented, fabricating in total 24 samples. The factor levels are given in Table \ref{DoE_multi}.
\begin{table}[!h]
  \centering
  \caption{Process parameters and their levels.}\label{DoE_multi}
  \begin{tabular}{llrrrr}
    \hline
    Process parameter& Units& L1& L2 &Lower &Upper \\
    \hline
    PVA in deionized water& wt\%& 9& 13 &8  &15  \\
    BTO particle size& nm& 100& 300 &80 &350 \\
    Nozzle diameter & mm& 0.8& 1.2 &0.5 &1.5 \\
    Printing speed & mm/sec& 10& 20 &8 &25 \\
    \hline
  \end{tabular}
\end{table}
Note that the factor ranges are not provided by \cite{Renteria_2019} and are hypothetical here for illustrative purpose. The samples were printed by a Printrbot Simple Metal 3D printing machine. A cylinder of 1 inch outer diameter, 0.5 inch inner diameter, and 0.5 inch height was selected as the standard sample for the study. See Section 2 of \cite{Renteria_2019} for more experimental and post-processing details. Experimental results can be found in Tables 4-7 in their supplementary material.

\subsubsection{Code Examples}
We aggregate the data into a file named ``MultiObj.csv'', with column names ``Binder\_amount'', ``Particle\_size'', ``Nozzle\_diameter'', ``Printing\_speed'', ``y\_1'' (for relative density), ``y\_2'' (for piezoelectric coefficient), ``y\_3'' (for dielectric permittivity), and ``y\_4'' (for dimensional accuracy). The goal in multi-objective optimization is to learn the Pareto front: the set of optimal trade-offs, where an improvement in one objective means deteriorating another objective. Our four objectives are the maximization of the relative density, maximization of the piezoelectric coefficient, maximization of the dielectric permittivity, and minimization of the angle of deformation. According to the summary table in Figure \ref{PKGcompare}, we exemplify via the following three libraries: \{mlrMBO, Trieste, BoTorch\}, due to they allowing multi-objective optimization and external objective evaluation. For the Python library Dragonfly, multi-objective optimization is currently not implemented in the ask-tell interface, and one has to define a callable function to do multi-objective optimization. Again, in all the following code examples, the outputs are shifted to have zero mean and then scaled to have unit variance.

\begin{figure}[!h]
  \centering
  \includegraphics[scale=0.85]{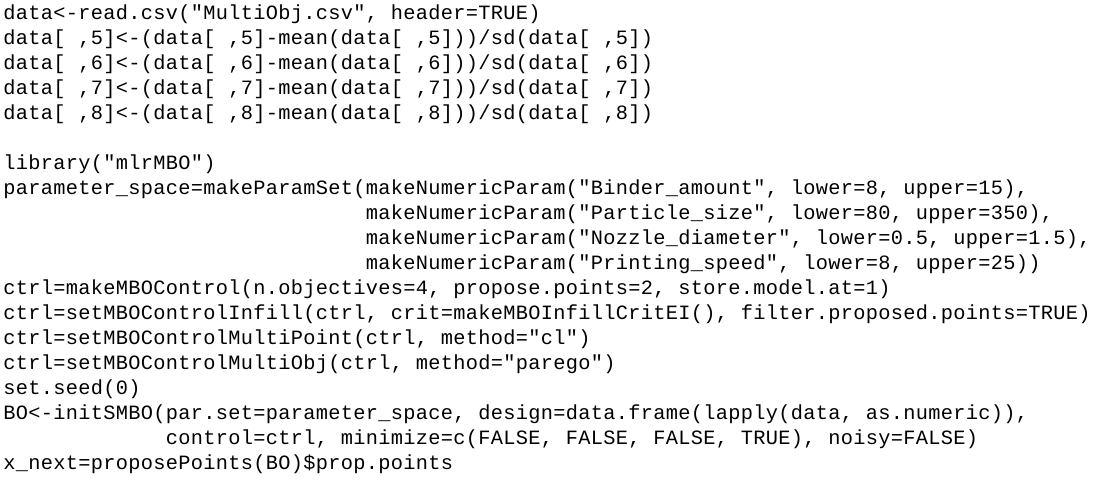}
  \caption{Code example for multi-objective optimization using mlrMBO.}\label{R_multi}
\end{figure}

\textbf{mlrMBO}: The code for multi-objective optimization with external evaluation is given in Figure \ref{R_multi}.
Compared with single-objective optimization, the main difference here is the additional function \textit{setMBOControlMultiObj}. There are quite a few options in \textit{setMBOControlMultiObj}, which require knowledge of multi-objective optimization. Here we simply accept all the default values. Optimization of noisy multi-objective functions is not supported at the moment. Setting of `final.method' and `final.evals' in the function \textit{makeMBOControl} is not supported for multi-objective optimization either. For illustrative purpose only, here we apply a different acquisition function and a different proposal of multiple infill points. The two suggested query points are (14.98, 80.00, 0.50, 13.62) and (14.99, 80.02, 1.49, 8.09).

\begin{figure}[!h]
  \centering
  \includegraphics[scale=0.85]{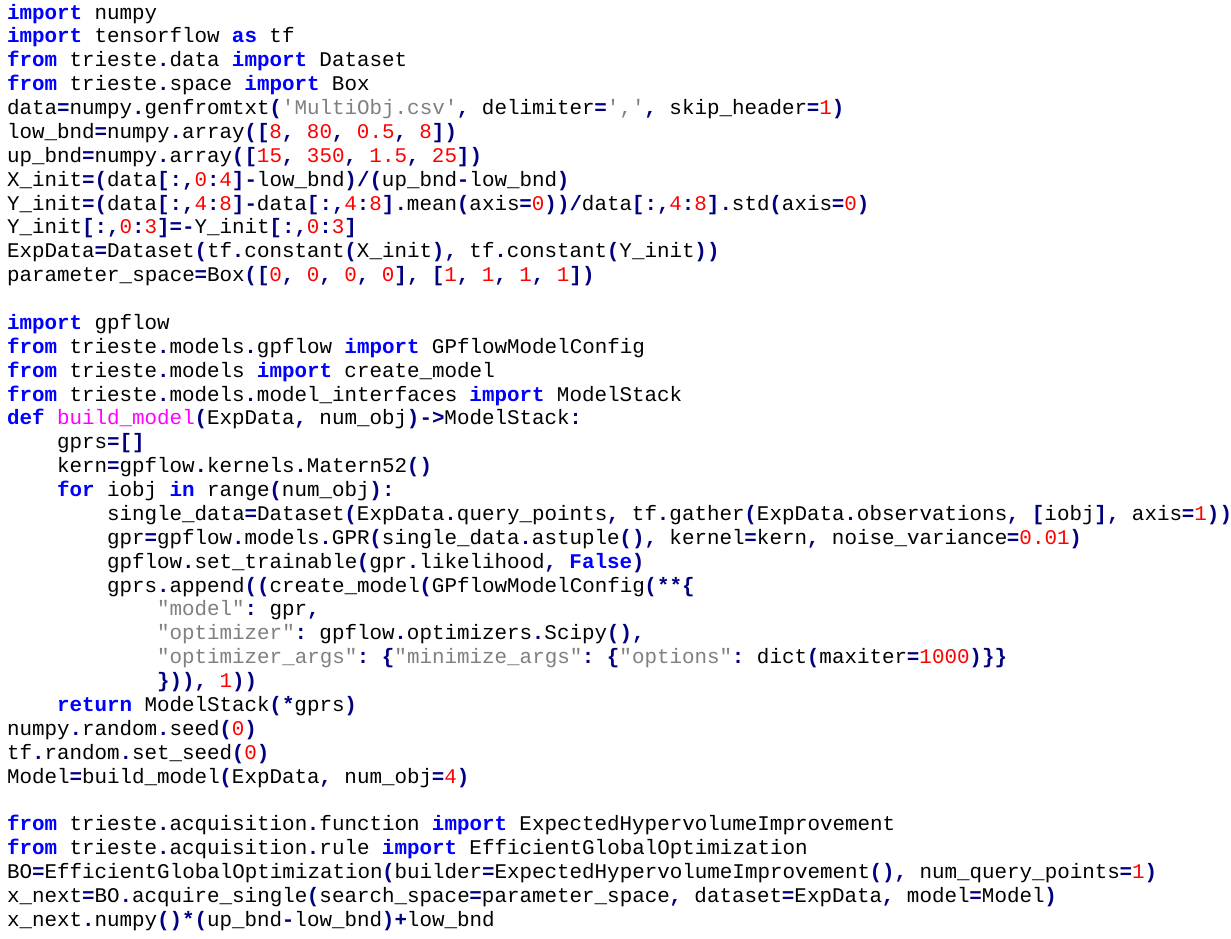}
  \caption{Code example for multi-objective optimization using Trieste.}\label{Trieste_multi}
\end{figure}

\textbf{Trieste}: Following the tutorials on the documentation site, we model the four objective functions individually with their own GP models. Figure \ref{Trieste_multi}
gives the code. The main changes are the function \textit{build\_model} and the new acquisition function. The function \textit{build\_model} simply builds a GP model for each objective and then applies the wrapper \textit{ModelStack} to stack these independent GPs into a single model working as a multi-output model. The acquisition function \textit{ExpectedHypervolumeImprovement} is for multi-objective optimization. Note that this acquisition function does not support batch optimization, and hence we set the argument `num\_query\_points' to be 1. The suggested query point is (12.11, 80.00, 0.82, 16.08).

\begin{figure}[!h]
  \centering
  \includegraphics[scale=0.85]{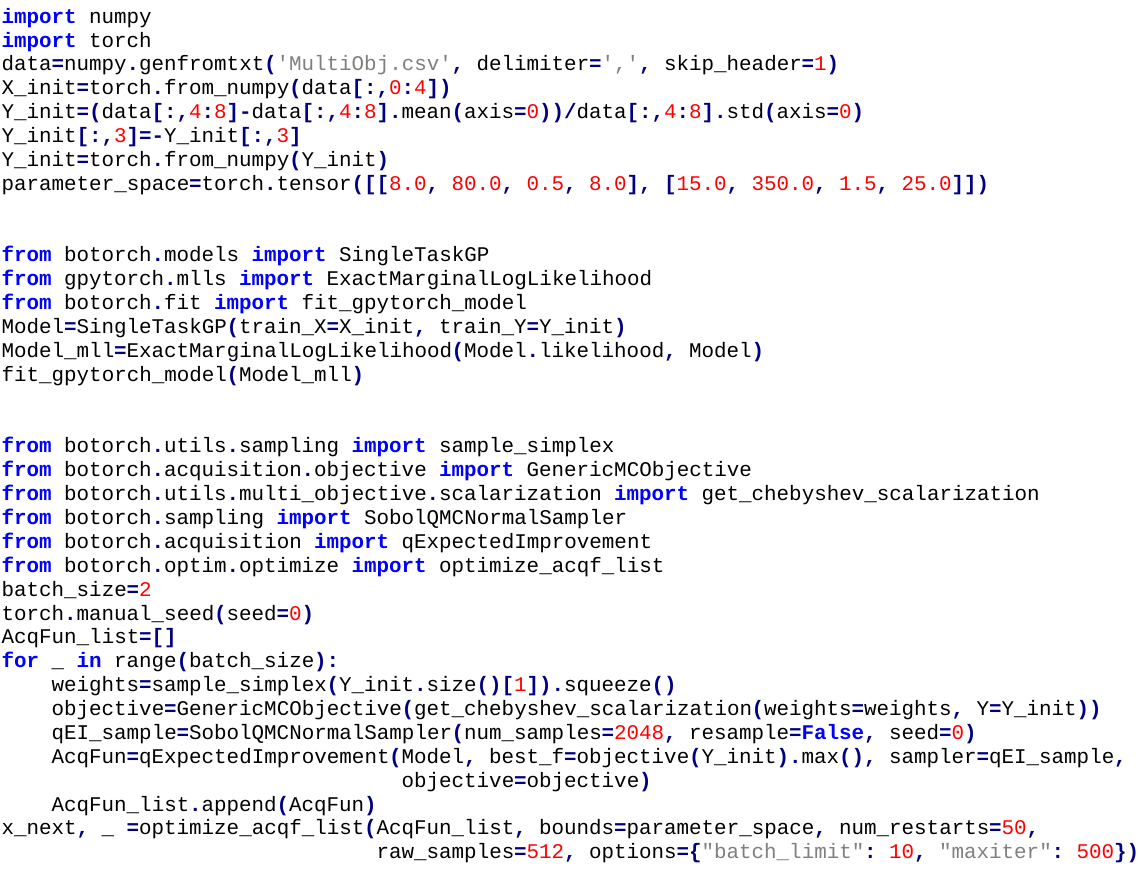}
  \caption{Code example for multi-objective optimization using BoTorch.}\label{BoTorch_multi}
\end{figure}

\textbf{BoTorch}: Likewise, we assume that the output features are conditionally independent given the input, and each output feature is noisy with a homoskedastic noise level. We apply \textit{SingleTaskGP} to fit a GP model for each objective. Figure \ref{BoTorch_multi}
gives the code example. In the batch evaluation setting (batch size $>$1), the code creates a list of \textit{qExpectedImprovement} acquisition functions, each with a different random scalarization weight, where the augmented chebyshev scalarization mehtod is applied to randomly scalarize the objectives. The \textit{optimize\_acqf\_list} method sequentially generates one candidate per acquisition function and conditions the next candidate (and acquisition function) on the previously selected pending candidates (see \cite{daulton2020differentiable} for details). The two suggested query points are (13.24, 99.91, 0.83, 20.13) and (9.52, 100.59, 0.93, 19.72).

From the functionality point of view, we have a clear winner: BoTorch, for multi-objective optimization with external evaluation. BoTorch is capable of performing batch multi-objective optimization on noisy data. By contrast, mlrMBO cannot deal with noisy data, and Trieste cannot do batch optimization.

\subsection{Optimization with Black-box Constraints}
\subsubsection{Background}
In the previous paste extrusion example, we attempted to optimize the piezoelectric coefficient metric and the relative density metric, along with another two metrics. However, for certain applications, we do not need to optimize the piezoelectric coefficient or relative density but to keep each of them within an acceptable range. In other words, we can re-formulate the four-objective problem into a constrained two-objective problem, where the constraints are that both the piezoelectric coefficient and the relative density need be large enough. Another AM example is fabricating NiTi products \citep{Mehrpouya20194691}, where we want to optimize over laser power, laser speed, and hatching space such that the recovery ratio (the ratio of the recoverable strain to the total strain) is as high as possible and that the austenite finish temperature is below a threshold. In both examples, we have two types of goals: one is to optimize certain metrics, and the other is to satisfy requirements on other metrics. In the nomenclature of optimization, we call such problems as constrained optimization problems.

For classical constrained optimization problems, the constraints can be formulated in analytical form. By contrast, a distinguishing feature of the above constrained problems is that the constraints are black-box: just like the objective function, the constraints are unknown and can only be evaluated pointwise via expensive queries to ``black boxes'' that may provide noise-corrupted values. More formally, we are interested in finding the global maximum of a scalar objective function $f(\mathbf{x})$ over some bounded domain, typically $\mathbf{x}\in \mathscr{X}$, subject to the non-positivity of a set of constraint functions $c_1$, ..., $c_k$. We write this as
\begin{equation}\label{con-BO}
\max\limits_{\mathbf{x}\in \mathscr{X}} f(\mathbf{x}),~~~\mbox{s.t.}~~~c_1(\mathbf{x})\leq0, \ldots, c_k(\mathbf{x})\leq0,
\end{equation}
where $f$ and $c_1$, ..., $c_k$ are all black-box functions. We seek to find a solution to Equation (\ref{con-BO}) with as few queries as possible.

We here use the data from \cite{Mehrpouya20194691} for fabricating NiTi shape memory alloys. A ProX 200 direct metal printer was used to fabricate parts on a NiTi substrate, and the fabrication chamber was purged continuously with Argon. The layer thickness was fixed at 30 $\mu m$, and the scanning strategy was bidirectional. 17 cylindrical samples with a diameter of 4.5 mm and a height of 10 mm were fabricated. Laser power, scanning speed and hatch spacing were varied among the samples, and the outputs of interest are transformation temperature and recovery ratio. Transformation temperatures were realized by a Perkin-Elmer DSC Pyris with a heating and cooling rate of 10$^{\circ}$C/min, and determined by tangent method from the DSC curve. For the mechanical testing, a 100 kN MTS Landmark servo-hydraulic was used for the compression test which was done by a $10^{-4}$/sec strain rate. For strain measurement, an MTS high-temperature extensometer was attached to the grips. The chamber temperature was kept at Af+ 10$^{\circ}$C for mechanical testing.

\subsubsection{Code Examples}
We aggregate the data into a file named ``BBcon.csv'', with column names ``Laser\_power'' (W), ``Scan\_speed'' (mm/s), ``Hatch\_spacing'' ($\mu m$), ``Recovery\_ratio'' (\%) and ``Austenite\_finish'' ($^{\circ}$C). Our objective is to find the optimal design, with as few experiments as possible, such that the recovery ratio is close to 100\%, and the Austenite finish temperature is lower than 10$^{\circ}$C. The libralies DiceOptim, laGP, Spearmint, GPflowOpt and BoTorch are all capable of dealing with black-box constraints. We here apply the R package DiceOptim and the Python package BoTorch, due to they allowing external objective evaluation.

\begin{figure}[!h]
  \centering
  \includegraphics[scale=0.85]{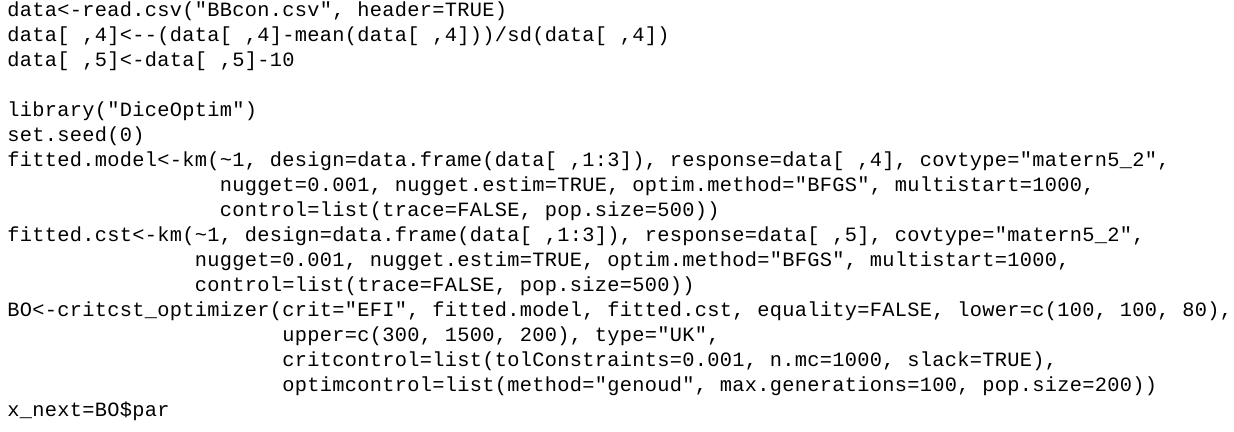}
  \caption{Code example for constrained optimization using DiceOptim.}\label{R_BBcon}
\end{figure}

\textbf{DiceOptim}: A code example is given in Figure \ref{R_BBcon}.
Since DiceOptim assumes a minimization problem, we flip the sign of the recovery ratio from positive to negative. We subtract all the temperature measurements by 10 to transform the constraint into the form of Equation (\ref{con-BO}). In the second block of code, we approximate both the objective function and the constraint function by GP models, where we add a small nugget effect to avoid numerical problems. For the optimization part, we select expected feasible improvement (EFI) as the acquisition function, as suggested by \cite{sym12101631}. DiceOptim supports equality and inequality black-box constraints. Running the code above, we will get the next query point (230.54, 190.60, 161.19). Note that, for a callable objective function, one can apply the function \textit{EGO.cst} for sequential constrained optimization and model re-estimation.

\begin{figure}[!h]
  \centering
  \includegraphics[scale=0.85]{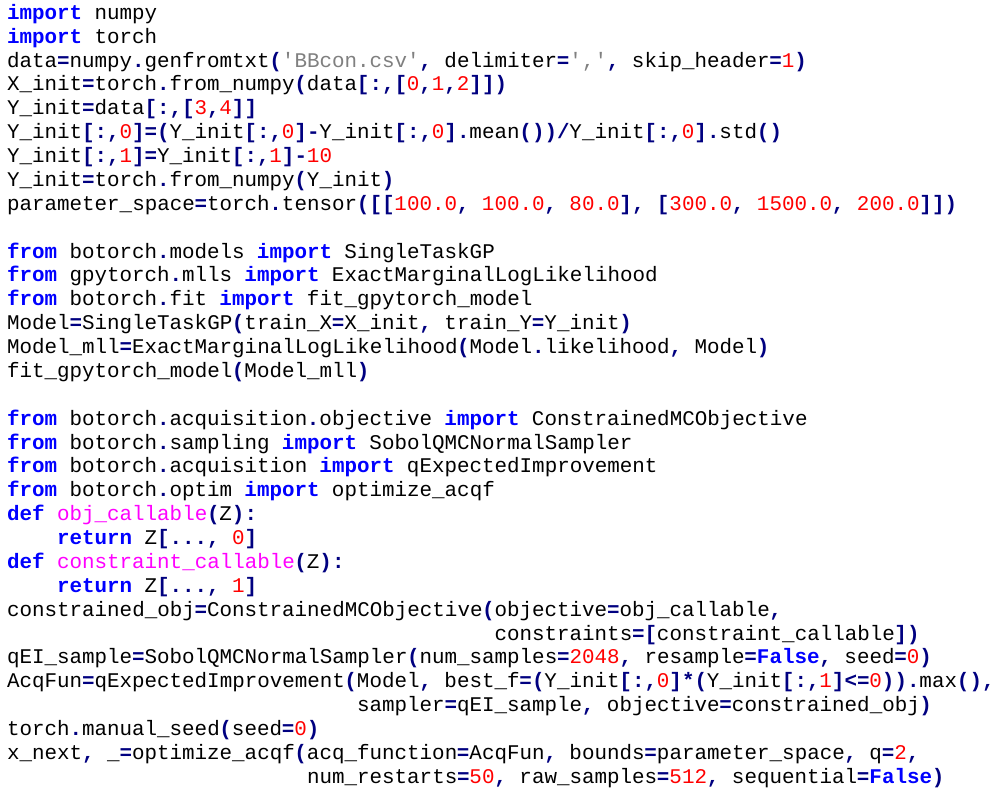}
  \caption{Code example for constrained optimization using BoTorch.}\label{BoTorch_con}
\end{figure}

\textbf{BoTorch}: The code example given in Figure \ref{BoTorch_con}
is similar to that for the batch optimization problem. The only difference is the additional function \textit{ConstrainedMCObjective} that computes the feasibility-weighted expected improvement in the objective. The two callable functions, \textit{obj\_callable} and \textit{constraint\_callable}, define a construct to index the objective function and constraint function. The idea here is similar to that of DiceOptim: fitting a cheaper statistical model to the underlying expensive-to-evaluate constraint function, just like the objective function is modeled by a surrogate model. Running the code above, we will get two query points (165.52, 1470.11, 156.68) and (100.06, 127.79, 114.50).

Compared with DiceOptim, AM practitioners may find it more difficult to understand the BoTorch code. However, after these three case studies, it is clear that BoTorch is the most flexible among all the libraries listed in Figure \ref{PKGcompare}. To conclude, we do not need worry too much about the performance/efficiency of a BO library. It is the functionality that dominates our choice.

\subsection{Preferential Output}
The performance of a printing job is usually characterized by one or multiple expensive-to-evaluate metrics: surface roughness, mechanical properties, dimensional accuracy and tolerances, etc. It may not be economical or efficient to collect the exact values of a performance metric. For example, optical measurement techniques such as focus variation, fringe projection, confocal laser scanning microscopy, and profilometers can be used to measure the surface roughness of fabricated products. However, different optical measuring methods produce (significantly) different values \citep{LAUNHARDT2016217}; from the optimization point of view, it is the ordering of the measurements (rather than the values of the measurements) that are of importance here. Mechanical-property tests (such as deformation property tests and failure property tests) are for quantifying mechanical properties like tensile strength, hardness, and fatigue strength. However, mechanical-property tests involve specialized test equipment, custom-fabricated fixtures, skilled test engineers, etc., and hence are commonly performed by commercial companies, which leads to a lengthy printing-testing cycle. For void or porosity analysis, X-ray computed tomography (micro-CT) is now the leading technique to analyze AM products non-destructively. However, the availability of micro-CTs and the knowledge of image analysis limit the application of micro-CT for porosity quantification. Again from the optimization point of view, we do not need that much information from micro-CT images but the information of which of two products has a lower porosity.

The idea behind collecting preference data instead of real-valued measurements is to accelerate and economize the DoE procedure, by replacing an advanced measuring technique with a simple one (or with a low-fidelity simulation model) and reaching a preferential judgement \citep{shahriari2015taking,gonzalez2017preferential,benavoli2020preferential}. Our output now is not a real value $y_i$ but a pairwise preference of the form $\mathbf{x}_i\succ \mathbf{x}_j$, meaning that the input configuration $\mathbf{x}_i$ is preferred over $\mathbf{x}_j$. We can create a list to store the preference data: $\mathbf{P}=\{(\mathbf{x}_i, \mathbf{x}_j), (\mathbf{x}_k, \mathbf{x}_m)\ldots\}$, where the order implies the preference. Alternatively, we can define a symmetric matrix $\mathbf{P}=[p_{ij}]_{n\times n}$ to represent the preference data:
\begin{align*}
p_{ij}=\left\{
         \begin{array}{ll}
           1, & \hbox{if } \mathbf{x}_i\succ \mathbf{x}_j;\\
           -1, & \hbox{if } \mathbf{x}_i\prec \mathbf{x}_j;\\
           0, & \hbox{if no preference between } \mathbf{x}_i \hbox{ and } \mathbf{x}_j.
         \end{array}
       \right.
\end{align*}
We want to find the optimum of the black-box function $f$ with minimal preferential information. Note that we are not against collecting real-valued measurements, and preferential BO is of advantage only when obtaining the preference $\mathbf{x}_i\succ \mathbf{x}_j$ is much cheaper than obtaining the real output $y_i$. In fact, BO can deal with mixed data where we have both preference data and real-valued measurements.

We here give a data-analysis example using the data from \cite{CAO2021107246}. The study investigated the influence of process parameters (laser power, scanning speed, and layer thickness) on surface roughness and dimensional accuracy. Experiments were conducted with a fiber laser operated in continuous mode with a wavelength of 1071 nm and a spot diameter of 50-80 $\mu$m. Surface roughness was measured with a surface-roughness tester (Mitutoyo  Surftest SJ-210) with 2.5 mm sampling length and 750 $\mu$m/s measuring speed. A laser scanning confocal microscope (Smartproof 5) was also applied to analyze the surface topography. The dimensions of the parts were measured using a digital vernier caliper with a precision of 10 $\mu$m. Taguchi L16 orthogonal array design was applied with three factor at four levels, and additional five random experiments were conducted, giving in total 21 initial experiments. To reduce randomness in experimental results, three cuboid samples with dimensions 45 mm $\times$ 5 mm $\times$ 1 mm were fabricated in each experiment with the same design. The lower and upper bounds for the three process parameters are \{180 W, 20 $\mu$m, 500 mm/s\} and \{240 W, 35 $\mu$m, 800 mm/s\}.

We copy the data into a file named ``SurfRough.csv'', with the original column names: ``Laser\_power'', ``Scanning\_speed'', ``Layer\_thickness'', ``Surface\_roughness'', ``Length'', ``Length\_dif'', ``Width'', ``Width\_dif'', ``Height'', and ``Height\_dif''. In this example, we focus only on the surface roughness, not the dimensional accuracy. We assume that the surface-roughness measurements are of variation, due to  aleatory uncertainty and/or epistemic uncertainty, and the error bound is 1 $\mu$m. That is, taking the first experiment as an example, the mean surface roughness could take any value in the interval [4.797, 6.797]. According to the 21 surface-roughness measurements, we can create a preference list $\mathbf{P}=\{(\mathbf{x}_1, \mathbf{x}_2), (\mathbf{x}_2, \mathbf{x}_3)\ldots\}$: we have $\mathbf{x}_1\succ \mathbf{x}_2$, $\mathbf{x}_2\succ \mathbf{x}_3$, and there is no preference between, e.g., $\mathbf{x}_1$ and $\mathbf{x}_6$. Given the preferential data $\mathbf{P}$ and a GP prior on the black-box function $f$, the posterior is a skew Gaussian process (SkewGP)  \citep{benavoli2020skew,benavoli2021arXiv}. Note that, different from the case of real-valued measurements where the surrogate model in the BO framework is GP, for the case of preferential output, the surrogate model in the BO framework is SkewGP. We therefore apply the SkewGP Python library\footnote{The library is available at \url{https://github.com/benavoli/SkewGP}.} on the preferential data $\mathbf{P}$ to get the next query point.

\begin{figure}[!h]
  \centering
  \includegraphics[scale=0.85]{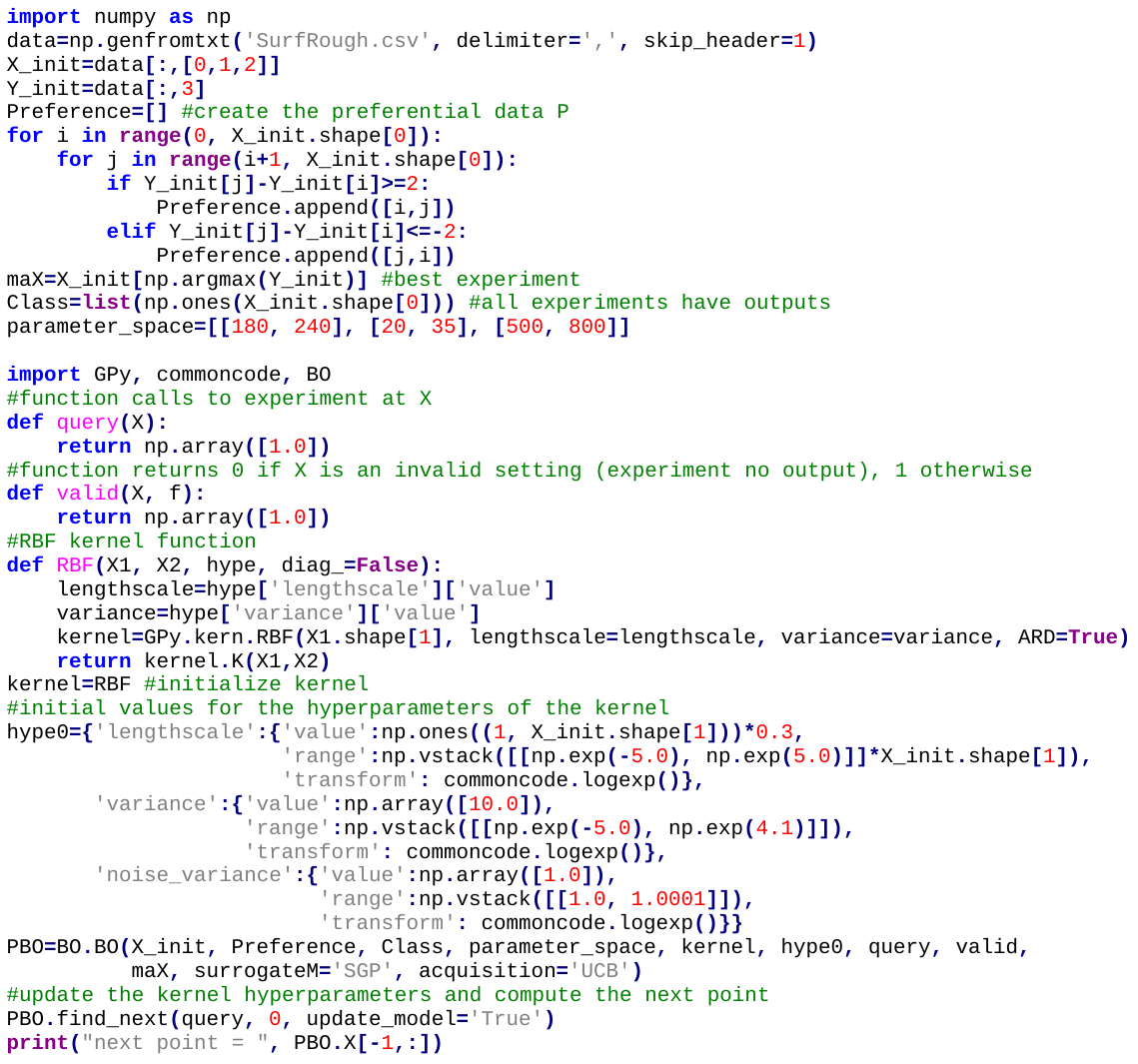}
  \caption{Code example for preferential BO using the SkewGP Python library.}\label{PBO_CodeExample}
\end{figure}

Running the code in Figure \ref{PBO_CodeExample},
we will get the next query point $(212.93,  28.89, 743.46)$. Here, we created the preferential data $\mathbf{P}$, a list, from the original data, while in practice the preferential data and the variable ``Class'' are generally read from external files. The SkewGP library can deal with outputs of mixed types. For example, if an experiment fails to print any product or prints a faulty product, then the experimental setting is invalid, and the output will be like $(\mathbf{x}_i, 0)$, where 0 means invalid (and 1 means valid); in other words, we will get the judgment of invalid, not any pairwise preference. Therefore, in our code example, we define the variable ``Class'', a vector of all 1's, to indicate that all experiments are valid.

\section{Conclusions}\label{Conc}
BO is a methodology for sample-efficient learning and optimization. By leveraging a probabilistic model, it is able to explore a large design space using only a small number of experimental trials. Our review in the field of AM reveals that BO, although indispensable for a competitive and digital AM, receives few applications until very recently. To facilitate and accelerate the adoption of BO by AM practitioners, we tabulated mature open-source BO packages concerning nine important practical considerations, serving as a guide on software selection. We also summarized the findings of existing comparative studies of BO packages, imparting more information on their relative efficiency. To help engineers dive right into coding their own problems, we provided code examples for several simple yet representative case studies, serving as a convenient one-stop reference. We wound up this work with an example of applying BO to new types of data, hoping to inspire creative applications and increase interest in BO. As a general outlook, there is a trend toward autonomous/digitalized AM, and we believe BO is a key technological element for future intelligent manufacturing. We have made updates to our code in response to changes made in certain Python packages, and you can find the latest version at https://github.com/GuangchenW/BayesianOptimization (last updated on September 19, 2023).

\section*{Acknowledgements}
This publication has emanated from research supported by a research grant from Science Foundation Ireland (SFI) under grant number 16/RC/3872 and is co-funded under the European Regional Development Fund. We want to thank Caitriona Ryan for discussions and suggestions.

\bibliographystyle{apalike}
\bibliography{RefSDoE}
\end{document}